\documentclass{article}
\usepackage[utf8]{inputenc}
\usepackage{graphicx}
\usepackage{cite}
\usepackage{amsmath,amssymb,amsfonts}
\usepackage{algorithmic}
\usepackage{graphicx}
\usepackage{textcomp}
\usepackage{xcolor}
\usepackage{booktabs} 
\usepackage{enumitem}
\usepackage{authblk}
\usepackage{url}
\usepackage{rotating}

\title{Connections between Relational Event Model and Inverse Reinforcement Learning for Characterizing Group Interaction Sequences}
\author{Congyu Wu\footnote{Correspondence: congyu.wu@austin.utexas.edu}}
\affil{University of Texas at Austin}

\date{October 2020}

\begin{document}

\maketitle

\begin{abstract}
In this paper we explore previously unidentified connections between \textit{relational event model} (REM) from the field of network science and \textit{inverse reinforcement learning} (IRL) from the field of machine learning with respect to their ability to characterize sequences of directed social interaction events in group settings. REM is a conventional approach to tackle such a problem whereas the application of IRL is a largely unbeaten path. We begin by examining the mathematical components of both REM and IRL and find straightforward analogies between the two methods as well as unique characteristics of the IRL approach. We demonstrate the special utility of IRL in characterizing group social interactions with an empirical experiment, in which we use IRL to infer individual behavioral preferences based on a sequence of directed communication events from a group of virtual-reality game players interacting and cooperating to accomplish a shared goal. Our comparison and experiment introduce fresh perspectives for social behavior analytics and help inspire new research opportunities at the nexus of social network analysis and machine learning.
\end{abstract}

\section{Introduction}\label{sec:intro}

From interpersonal relations to international politics, human social systems are replete with sequences of directed social interaction events, or dyadic actions \cite{heise1997frame}. Person A sending person B a text message is a dyadic action. Country A providing financial aid for country B is also a dyadic action. Intuitively, these dyadic actions can be defined by their initiator (from whom), receiver (to whom), nature (what they were), and time (when they happened). If we observe a group of ``actors" initiating and receiving dyadic actions for a period of time, the resulting data would be a sequence of events with the above properties specified. Such data have been used as key evidence for understanding social systems in various applications such as workplace organizations and online classrooms \cite{wu2008mining, vu2015relational, joksimovic2016translating}. With the uptake of mobile sensing technology, we are increasingly able to obtain fine-grained, time-stamped dyadic action sequences among groups of mobile technology users interacting in natural settings, especially face-to-face communication \cite{olguin2008sensible}. 

To uncover the driving factors of group dyadic actions and potentially use them to predict future actions, a social network model called \textit{relational event model} (REM) was proposed \cite{butts2008relational} and remains a popular tool to analyze group dyadic action sequence data \cite{vu2015relational}\cite{joksimovic2016translating}. The premise of REM is that the likelihood of a particular sequence of group dyadic actions depends on a set of latent features and their respective degree of importance that are intrinsically possessed by the group regarding the type of dyadic action. For example, if a group of people place high importance on the reciprocation of actions (e.g., physical attacks that justify retaliation), then many back-and-forth dyadic actions should be observed; whereas if a social circle likes to ``pay it forward'' (e.g., circulating gossips), the dyadic actions among the members tend to form directed chains. Thus, given a pre-specified set of underlying features and their respective importance values, some dyadic action sequences are more likely to emerge than others. By fitting an REM model to collected group dyadic action sequence data, one can learn a set of importance values of the specified underlying features that maximize the likelihood of the observed sequence of dyadic actions. These learned importance values constitute valuable insight into how a group of individuals interact and form a social system.

Now we look to the field of machine learning. \textit{Reinforcement learning} (RL) is a type of machine learning problem which hypothesizes that the behavior of an intelligent agent is driven by its desire to accumulate higher \textit{reward} when interacting with its broadly construed environment. By taking different actions, an agent transitions to different states of the environment, resulting in different rewards at each transition; the actions that result in higher rewards would be favored in the future and those in lower rewards avoided. An RL algorithm assumes knowledge of such reward values and aims to find an optimal behavioral strategy, or policy, that would achieve the most reward over time. Its inverse problem, namely \textit{inverse reinforcement learning} (IRL) \cite{ng2000algorithms, abbeel2004apprenticeship, ziebart2008maximum, ramachandran2007bayesian} on the other hand, takes the observed sequence of states and actions of an agent as input and seeks to solve for the reward that must have driven the actions taken. In this sense, there exists a resemblance between IRL and REM, because both approaches take as input an observed event trajectory, both assume the existence of certain underlying measures that are affected by taking different actions and thus govern how different actions are sequentially chosen, and both try to find the values of these measures that can best explain the observed trajectory of events. 

This resemblance motivates us to explore two research questions. On a theoretical level, we want to inspect how REM and IRL are connected mathematically. We discuss technical details of REM and IRL in Sections \ref{sec:rem} and \ref{sec:irl}, examine the analogies and distinctions in their mechanisms in Section \ref{sec:connections}, and find many mathematical components of the two equivalent. We outline several unique aspects of REM and IRL that do not match in Section \ref{sec:unique}, which call for further research and potential new development. On a practical level, we wonder in what ways an IRL approach, which has not conventionally been utilized in social interaction modeling, would enrich the existing social network models such as REM to characterize group social interaction dynamics, especially given their differences in technical capability deliberated in the theoretical comparison. The modeling perspective of REM is a bird-eye view of the whole group thus making it unable to straightforwardly specify the nuances in the behavioral tendency of different individuals in a group. When a group interaction scenario involves different individuals with distinct roles and behavioral tendencies (e.g., team leaders and team members), understanding group behavior only from its entirety may become insufficient and the need to model individual behavior arises. We propose that IRL can address this limitation and demonstrate in Section \ref{sec:experiment} with real-world data how IRL can be used to model sequences of directed social interaction events in group settings. We showcase the unique utility of an IRL approach in characterizing individual behavior preferences. 

To the best of our knowledge, this paper is the first to (1) identify and explore the theoretical connection between relational event model and inverse reinforcement learning in modeling dynamics of directed social interaction events in groups, and (2) apply and show potential value of inverse reinforcement learning from the perspective of individual actors using data for which a network modeling approach is conventionally employed. 

\section{Related Work}
\subsection{Relational Event Model}\label{sec:rem}

\begin{figure}
    \centering
    \includegraphics[width = 0.8\columnwidth]{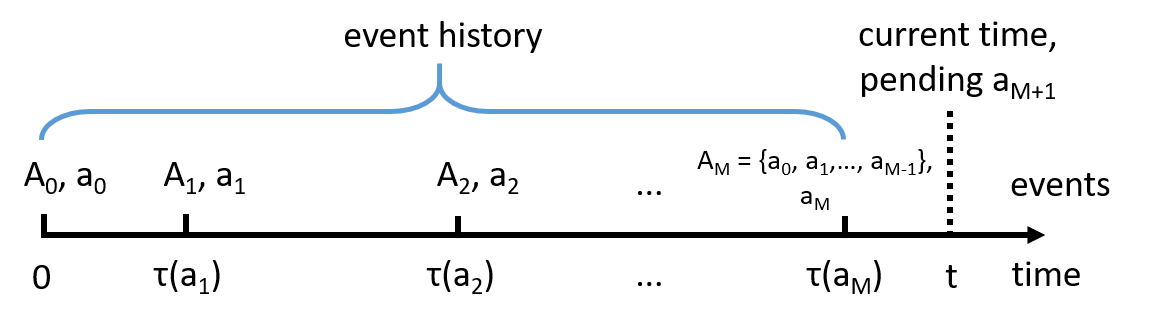}
    \caption{An illustration of timestamped relational event history, the input data for REM}   
    \label{fig:events}
\end{figure}

Relational event model is capable of taking full advantage of time-stamped or at least time-ordered group dyadic action data which are ubiquitous in human social systems. Full details can be found in its original paper \cite{butts2008relational} and we review the key building blocks here. First and foremost, a dyadic action or directed social interaction event $a \in \mathbb{A}$, where $\mathbb{A}$ is the set of all possible actions, is fully characterized by five elements: (1) a sender $s(a) \in S$, where $S$ is the set of senders; (2) a receiver $r(a) \in R$, where $R$ is the set of receivers; (3) an action type $c(a) \in C$, where $C$ is the set of action types; (4) a timestamp $\tau(a)$, and; (5) descriptive covariate(s) $X_a$, which may or may not be specified. At a given time $t$, we have observed $A_t$, a time-stamped (thus time-ordered) history of $M$ dyadic actions $a_1, a_2, ..., a_i, ..., a_M$ ($a_M$ is the most recent) and we denote $a_0$ as a place-holder for the beginning point of the observed sequence with its timestamp $\tau(a_0) = 0$. The event history is illustrated in Figure \ref{fig:events}.

To specify the probability density of such an event history, REM resorts to survival and hazard functions \cite{miller2011survival}. This is a fitting approach because an event history clearly consists of multiple eventless periods between discrete dyadic actions scattered over time. A survival function $S_a(t)$ expresses the probability of an action $a$ not happening over duration $0 \mathtt{\sim} t$ while a hazard function $h_a(t) =\frac{\partial [1-S_a(t)]}{\partial t}/S_a(t)$ quantifies the propensity that an action $a$ did occur when some action was to happen. As such, the probability of an event history can be written as a product of multiple survival functions and hazard functions as Equation 1. 

\begin{align}
    p(A_t) & = \prod_{i=1}^M[h_{a_i}(\tau(a_i))\prod_{a\in\mathbb{A}}S_{a}(\tau(a_i)-\tau(a_{i-1}))] \notag \\
    &\qquad \times\prod_{a\in\mathbb{A}}S_{a}(t-\tau(a_M)) \label{eq:ergm_prod}
\end{align}
\begin{align}
    h_{a_i}(\tau(a_i)) & = \lambda(a_i, A_{\tau(a_{i-1})}, X_{a_i})  \\
    & = \exp[\theta^{T}u(a_i, A_{\tau(a_{i-1})}, X_{a_i})] \label{eq:rate}
\end{align}

Based on the intuition that actions with different senders, receivers, types, and other descriptive covariates should have different likelihood of happening given preceding events, REM specifies the hazard $h$ as a rate function $\lambda$ (Equation 2) that is equal to the exponential of the weighted sum of a vector of \textit{sufficient statistics} $u(a_i, A_{\tau(a_{i-1})}, X_{a_i})$ (Equation 3). The sufficient statistics featurize the trajectory of past events leading up to the current action. $\theta$ is a coefficient vector associated with the sufficient statistics quantifying their importance. The rate function governs the distribution of probability associated with each action given the past event history. In cases where the temporal order of the events is available but exact timestamps are not, the probability of an event history realizing can be equivalently written as Equation 4 (proof available \cite{butts2008relational}), which relaxes the requirement for timestamped data.

\begin{align}
    p(A_t) = \prod_{i=1}^M[\frac{\lambda(a_i,A_{\tau(a_{i-1})},X_{a_i})}{\sum_{a'\in\mathbb{A}}\lambda(a',A_{\tau(a_{i-1})},X_{a'})}] \label{eq:ergm_ordinal}
\end{align}

By expressing the hazard function in terms of linear combinations of sufficient statistics, REM allows not only straightforward parameterization of certain realizations of event trajectory but also straightforward inference procedure through likelihood-based methods (e.g., maximum likelihood estimation) to learn the values of coefficients $\theta$ that best fit real data. We find applications of REM in diverse domains to understand human social behavior such as team processes \cite{leenders2016once}\cite{pilny2016illustration}, friendship \cite{foucault2014dynamic}, and within education \cite{vu2015relational}\cite{joksimovic2016translating} and health care \cite{vu2017relational} settings. The group interaction dynamics characterized by REM are anticipated to be useful for predicting individual and group outcomes such as task performance \cite{pilny2016illustration}. 

\subsection{Inverse Reinforcement Learning}\label{sec:irl}

The life of every living creature naturally involves observing changing environment and acting upon it in a way favorable for their survival and prosperity. In artificial intelligence applications where we train a computer to complete human tasks (e.g., playing chess), we also require it to be able to make ``good moves" given a situation (e.g., positions of self and opponent pieces) that are conducive to favorable outcomes (e.g., winning). As such, questions arise as to how a human or machine should choose actions based on the observed environment, which can change as a result of previous actions taken, to achieve particular goals over time. 

\begin{figure}
    \centering
    \includegraphics[width = 0.8 \columnwidth]{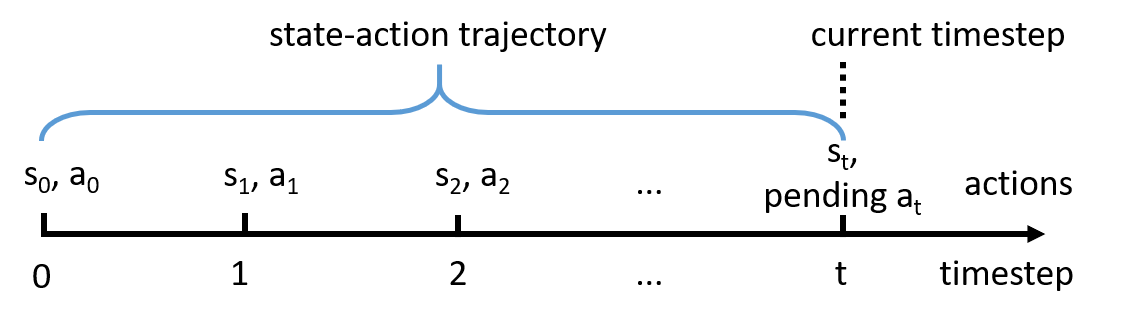}
    \caption{An illustration of state-action trajectory, the input data for IRL}   
    \label{fig:sa_trajectory}
\end{figure}

Reinforcement learning is a method aimed at officially answering these questions. In its basic setup, an agent finds itself in a \textit{state} $s$ at each time step $t$, which characterizes the context in which the agent is situated, and takes an \textit{action} $a$, which alters the context and thus brings the agent to a new state $s'$. Both $s$ and $s'$ belong in a \textit{state space} $S$, the set of all possible states, and all possible actions form an \textit{action space} $A$. State transition is governed by transition probabilities $P_{a}(s,s')$, specifying the probability that the agent will arrive in state $s'$ after taking action $a$ while in state $s$. Upon arriving in the new state $s'$, the agent receives a \textit{reward} (or \textit{reward function}) $R(s)$, which quantifies the desirability of being in the new state; a reward value is associated with each state in the state space $S$. The agent interacts with the environment by repeating the state-action cycle and receiving and accumulating reward in the meantime. The way the agent behaves follows a \textit{policy} $\pi(s) = P(a|s)$, governing the probability with which the agent is to take an action $a \in A$ when in state $s$. A deterministic policy specifies one action to take given a state whereas a stochastic one provides a probabilistic distribution over multiple actions. Following different policies, an agent will take actions and visit states differently, thus achieving different amount of accumulated reward over time. A \textit{value} (or \textit{value function}) $V^{\pi}(s_0) = \sum_{t=0}^{\infty}\gamma^{t}R(s_t) $ is defined for each state as the discounted accumulated reward when the agent starts from state $s_0$ and behaves onward following policy $\pi$. $\gamma$ is a discount factor converting a reward received later on into its current value, indicating a preference for long- or near-term reward seeking. The core problem of reinforcement learning is to find an optimal policy $\pi^*$ for the agent that maximizes its accumulated reward regardless of starting state, given knowledge of the states, actions, and rewards. This optimization problem can be solved by dynamic programming algorithms and has extremely broad applications in artificial intelligence. 

\begin{align}
    \pi^* = argmax_{\pi} V^{\pi}(s_0) = argmax_{\pi} \sum_{t=0}^{\infty} \gamma^{t}R(s_t), \forall s_0 \label{eq:argmax_value}
\end{align}

Reinforcement learning assumes knowledge of the reward and aims to solve for an optimal policy to serve as a behavioral guidance for the agent; for example, to direct a robot to successfully navigate a labyrinth or win a chess game against human opponents. However, in some behavioral cases (e.g., driving a car), we don't know or cannot define the reward straightforwardly (e.g., what is ``driving well"); instead, we can define the state and action spaces, observe an example agent's behavior, and want to find the underlying reward that are mostly likely to have guided the agent to behave the way it does. Such recovered reward can be further used to train other agents or simply to understand the example agent's behavioral pattern or strategy. This problem constitutes \textit{inverse reinforcement learning} (IRL), for which the central task is to solve for the reward given observations of an agent's state-action trajectory $O$ (as illustrated in Figure \ref{fig:sa_trajectory}).

Different IRL algorithms have been proposed to use different optimization strategies procedures to find reward given state-action trajectories. A mechanism adopted by several IRL algorithms \cite{ng2000algorithms, abbeel2004apprenticeship, ziebart2008maximum} is to characterize a state $s$ using a feature vector $f(s)$ and express the corresponding reward $R(s)$ as a linear combination of the features with the weight vector $\theta$ indicating the importance of a feature: $R(s) = \theta^{\top}f(s)$. With reward $R(s)$ decomposed into state features, the value of a state under a policy $V^{\pi}(s)$ can be decomposed into featurized state values by the same set of weights $\theta$. Some optimization procedures \cite{ng2000algorithms}\cite{abbeel2004apprenticeship} rely on these featurized state values by iteratively updating weights $\theta$ and the corresponding optimal policy $\pi^*$ until the resulting featurized state values are very close to those of the example agent's state-action trajectories. Other algorithms \cite{ziebart2008maximum}\cite{ramachandran2007bayesian} on the other hand, adopt a probabilistic perspective and aim to find the reward that maximizes the probability of the observed trajectories through procedures similar to maximum likelihood estimation. 

Once the reward function is learned, an optimal policy can be computed to generate agents that behave similarly to the demonstrated behavior of the example agent. We find applications of IRL in many motion imitation tasks (e.g., training drones or self-driving cars); however, using IRL to solve social interaction modeling problems is still a largely unbeaten path. One related work \cite{murray2017markov} showed promising results using Markov reward model, a Markov decision process based approach to learn favorable scenarios within group interactions.  

\section{Research Questions}

As introduced in Section \ref{sec:intro}, we ask the following two research questions. We address RQ1 by examining and comparing the mathematical components of IRL and REM and RQ2 using a demonstrative experiment implementing an IRL algorithm on real-world group interaction data that was modeled by REM in previous work.

\begin{itemize}
    \item \noindent\textbf{[RQ1]}: In what ways are the mathematical components of IRL and REM similar and different to one another?
    \item \noindent\textbf{[RQ2]}: In what ways can IRL enhance the characterization of directed social interaction events in groups that is conventionally done using network models such as REM? 
\end{itemize}

\section{Theoretical Analysis}
\subsection{Analogies}\label{sec:connections}

REM and IRL originated independently from two different fields; however, both posit \textit{actions} as a central concept and theorize that the tendency of a subsequent action depends on a current situation resulting from recently realized history. Such similarity inspires us to examine parts of REM and IRL that are analogous. In Table \ref{tab:matching} we list elements of the two methods that we find equivalent to one another. 

\begin{sidewaystable}[]
\centering
\caption{Equivalent components of REM and IRL algorithms}
\label{tab:matching}
\begin{tabular}{@{ }llll@{ }}
\toprule
  \multicolumn{2}{c}{REM}  & \multicolumn{2}{c}{IRL}                \\ \midrule
  Group (of $N$ actors)   &    & Agent      &                   \\ 
  Event history & $A_i = A_{\tau(a_{i-1})}$ & State & $s_i$  \\ 
  Dyadic action & $a_i$        & Action & $a_i$  \\ 
  Newly realized event history & $A_{i+1} = A_{\tau(a_{i})} = \{a_i, A_{\tau(a_{i-1})}\}$ & New state & $s_{i+1}$\\
  Trajectory of $M$ realized histories & $\{A_1, ...,  A_M \}$ & State trajectory of length $T$ & $\{s_1, ..., s_T \}$\\
  Rate/hazard function & $\lambda(A_i) = \exp[\theta^{\top} u(A_i)]$ & Reward function & $R(s_i) = \theta^{\top} f_{s_i}$\\
  Sufficient statistic & $u(A_i)$    & State feature & $f(s_i)$                 \\
  Coefficient & $\theta$        & Reward weight & $\theta$       \\\bottomrule
\end{tabular}
\end{sidewaystable}

First, the group of actors studied in REM is equivalent to an IRL agent. Although it is reasonable and somewhat more intuitive to conceive an individual sender in REM as an IRL agent (on which we will elaborate in Section \ref{subsec:unique_agent}), the group in REM as a whole is equivalent to the IRL agent because (1) under the setting of REM it is the group as a whole rather than individual senders that are the observer of their own event history, which encompasses behaviors of all senders, and; (2) the goal of REM is to infer underlying group-level, rather than sender-specific, characteristics that drive group social interaction dynamics.

Second, we find directly matching notions in REM for \textit{action} and \textit{state} in IRL. Straightforwardly, a relational event or dyadic action $a$ in REM maps to the action $a$ in IRL. Action space $\mathbb{A}$ in REM is determined by the number of senders $|S|$, the number of receivers $|R|$, and the number of action types $|C|$. Assuming all dyadic actions in a group are legal and no self-directed actions are permitted, the action space size would be $N(N-1)\times|C|$, where $N$ is the total number of nodes in the group. In REM, the notion of an IRL \textit{state} is effectively fulfilled by a group's past relational event history, as it ``creates the context for present action, forming differential propensities for relational events to occur" \cite{butts2008relational}. A newly taken dyadic action $a_i$ directly updates the past event history $A_\tau(a_{i-1})$ of the group at the time $a_i$ is taken, as the newly taken dyadic action becomes the most recent event in event history, and thus places the group in a new state. The new state $s'$ is simply a concatenation of the old state and the action taken upon it: $A_\tau(a_i) = \{a_i, A_\tau(a_{i-1})\}$. In other words, REM is intended for a Markov decision process (MDP) with \textit{history} as \textit{state}. This property assumes group interaction process to be (1) a deterministic MDP, because once a new event happens, the event with 100\% probability gets placed on top of past event history, and (2) an MDP with non-revisitable states, as event history always grows (assuming infinite memory) and once a dyadic action is realized, none other could happen. As IRL algorithms require state-action trajectories as input, and we find the nature of past event history in REM equivalent to an RL state, we find that the input data for REM is also suitable for IRL. 

Third, the core of both the REM and IRL approaches lies in the mechanism to determine what actions are more likely or favorable for the group/agent to take given a context. This notion of context takes the form of event trajectory $A$ in REM and state $s$ in IRL. Both methods quantify the desirability of a to-be-realized context by a function: in REM it is the \textit{rate function} $\lambda$ while in IRL it is the reward function $R(s)$, or simply reward $R$ when not stressing the mapping between a state $s$ and the desirability of being in it. Given a current state, actions that result in transition to a subsequent state of a higher rate value in REM or reward value in IRL are more likely chosen than those of lower values. To specify the different factors that contribute to the desirability of a to-be-realized context, REM decomposes the rate function $\lambda$ linearly into a weighted sum of sufficient statistics $u(A)$. This mechanism is shared by multiple IRL algorithms such as Apprenticeship Learning IRL \cite{abbeel2004apprenticeship} and Maximum Entropy IRL \cite{ziebart2008maximum}. In these IRL algorithms, the reward of being in a state is also decomposed linearly, as a weighted sum of state features $f(s)$. Both the sufficient statistics in REM and the reward features in these IRL algorithms are the \textit{de facto} descriptors of a context and need to be pre-specified and measured by the researcher, which are analogous to the explanatory variables or predictors in supervised learning. Once these descriptors are selected, the objective of both REM and the IRL algorithms is to find the value of their corresponding weights $\theta$ via optimization techniques. These inferred $\theta$ values, together with the observed values of the sufficient statistics or state features, allow straightforward calculation (by taking a weighted sum) of the desirability to the group/agent of a specific event trajectory/state, thus revealing the group/agent's behavioral tendencies. Researchers have indeed used the inferred $\theta$ values as input for further modeling and prediction tasks such as behavioral clustering \cite{lee2017agent} and team performance prediction \cite{lomi2011some}. 

\begin{align}
    P(O) & = P[(s_1, a_1), (s_2, a_2), ..., (s_k, a_k)]\label{eq:birl_whole} \\
    & = \prod_{i=1}^{k} P(s_i, a_i)\label{eq:birl_prod} \\ 
    & = \prod_{i=1}^{k} \frac{\exp{[R(s_{i})]}}{\sum_{s' \in S}{\exp{[R(s')]}}}\label{eq:birl_exponent}
\end{align}

Last but not least, we discover that the objective function used in the optimization procedure of REM to find the most fitting rate function and weights is equivalent to that of another major IRL algorithm called Bayesian IRL \cite{ramachandran2007bayesian}. Generally, despite the commonality in the setup of action, state, and the goal of inferring reward, different IRL algorithms use different optimization procedures. Bayesian IRL, specifically, adopts a strategy identical to that of REM which is to express the likelihood of an action being taken at each step in the trajectory (as opposed to expressing that of entire trajectories as in Maximum Entropy IRL), and then maximize the product of step-wise likelihoods as the likelihood of the demonstrated event trajectory (Equation \ref{eq:ergm_ordinal}). The Bayesian IRL algorithm first expresses the likelihood of an state-action trajectory $O$ (Equation \ref{eq:birl_whole}) as the product of likelihood of state-action pair at each timestep (Equation \ref{eq:birl_prod}) based on the independence assumption. Then it models the likelihood of each state-action pair $(s_i, a_i)$ as proportional to the exponential of the value $V$ of taking action $a_i$ while being in $s_i$. This value is the accumulated reward expected from currently being in that state and all future states (see Equation \ref{eq:argmax_value} in Section \ref{sec:irl}), and when we consider the future discount factor $\gamma$ to be zero, is equal to the reward of state $s_i$, hence Equation \ref{eq:birl_exponent}. Equation \ref{eq:birl_exponent} turns out to have the identical form to that of Equation \ref{eq:ergm_ordinal} of REM.

\subsection{Distinctions}\label{sec:unique}

\subsubsection{Agent identity}\label{subsec:unique_agent}

What constitutes an \textit{agent} is the first design choice when formulating a Markov decision process for IRL. The perspective of REM is a bird-eye view of a whole group of $N$ actors: legal actions between \textit{all} pairs of actors are considered and ranked by their propensity to be taken. Learned coefficients of the sufficient statistics also represent the interaction dynamics of the whole group as opposed to individual actors. Therefore, the counterpart IRL problem should consider the group as the agent. However, IRL is a more generic approach that can comfortably treat an individual actor in the group as agent and other actors and their behaviors as environment if we construct the MDP accordingly. This way, the modeling perspective becomes truly egocentric. Compared to REM, IRL affords higher freedom in choosing what subset of a social system to be the agent and what subset to be the environment, contingent upon the problem at hand. We anticipate IRL to be useful for understanding group interactions in the international politics domain (in which previous work focused on network science approaches \cite{hafner2009network}) where behaviors of individual countries may be of higher interest than collective dynamics of multiple countries.

\subsubsection{State space}\label{subsec:unique_states}

As shown in Equation \ref{eq:ergm_ordinal}, the entire event history is retained to fit an REM. However, the sufficient statistics of REM require memory of different lengths of history, often much shorter than the entire history since $t_0$. For example, to calculate the statistic \textit{reciprocity} defined as whether a current action is of a reciprocal nature, one would only need information about the most recent event, and if the sender and the recipient of the current dyadic action correspond to the recipient and the sender of the most recent event respectively, the feature is assigned value 1 and all otherwise 0. However, in cases like feature \textit{inertia} defined as the degree to which a current action is repeated in the past, one would need to look up the current action in the entire event history and calculate the frequency. This difference does not fundamentally affect REM modeling; however, it reflects greatly in the construction of state space when we formulate an MDP for IRL. REM amounts to an IRL problem with a state space consisting of slices of non-revisitable, potentially-realizable event histories of all possible lengths, which may result in extremely large state spaces. However, if we truncate the slices of event history populating the state space to a fixed number of recent actions, the states become revisitable and the state space considerably smaller. Depending on the sufficient statistics/features we choose, and out of the intuition that older events may not matter as much as more recent ones, we may not need to retain the entire event history all the time; the truncation operation may be a viable, even advisable choice when building IRL procedure for group interaction modeling problems. 

In an MDP with recently realized action histories as states, the size of state space is eventually determined by the number of past actions retained as part of a state and the number of possible actions, the latter of which is in turn determined by the number of action-capable actors and event types. As the size of state space increases at an exponential rate as do the number of possible actions and the number of recent actions, over-large state spaces are a pressing issue to resolve and a thinly veiled curse of dimensionality. Countermeasure strategies targeting over-large state spaces are of primary interest and mainly fall into two categories: (1) state aggregation through clustering actors, actions, and state features to directly reduce state space size; (2) value function approximation to generalize from seen states to unseen ones.

\subsubsection{Model assumptions}\label{subsec:unique_assumption}

Besides agent identity and state space, several model assumptions also show distinctions of IRL that speaks to its flexibity compared to REM. First, REM explicitly assumes the absence of forward looking \cite{butts2008relational} whereas IRL is naturally equipped with the discount factor $\gamma$ to handle future states and how they are reflected in the current decision making. REM essentially amounts to IRL models with $\gamma = 0$.

Also, IRL and REM have different assumptions on the stochasticity of action choosing. The current machinery of REM has one rule of choosing actions: given a past event history, a rate function value is computed for each possible next action and the probability of an action being chosen as the next action is modeled to be proportional to its rate function value. In this way an action with a higher rate function value has a higher probability to be chosen. This operation amounts to the Thompson sampling, or probability matching strategy \cite{scott2010modern} in the multi-armed bandit problem. In IRL, multiple action choosing strategies have been proposed. A popular one is $\epsilon$-greedy strategy, whereby an agent chooses the action with the highest reward with probability $1-\epsilon$ and randomly chooses among all possible actions with probability $\epsilon$. Different probabilistic rules for action choosing could be incorporated into REM as well. We suspect that such incorporation is beneficial in behavioral modeling as different real-world social systems fit different assumptions of action choosing rules to different degrees.  

Compared to IRL, REM makes up its lower flexibility with its ability to also model timestamped event sequence, thanks to its survival analysis mechanism (concretely, Equation \ref{eq:ergm_prod}). It is not within the current machinery of IRL to take advantage of continuous timestamps as the states are treated as discrete. The philosophical trade-off between REM and IRL is that REM is a more ``ad-hoc", special purpose model and theoretical framework without the horns and whistles of IRL whereas IRL serves a wider range of purposes but may feel somewhat unnatural dealing with specific issues when applied to group interaction modeling problems.

\section{Demonstrative Experiment}\label{sec:experiment}

\begin{figure} 
    \centering
    \includegraphics[width = 0.8\columnwidth]{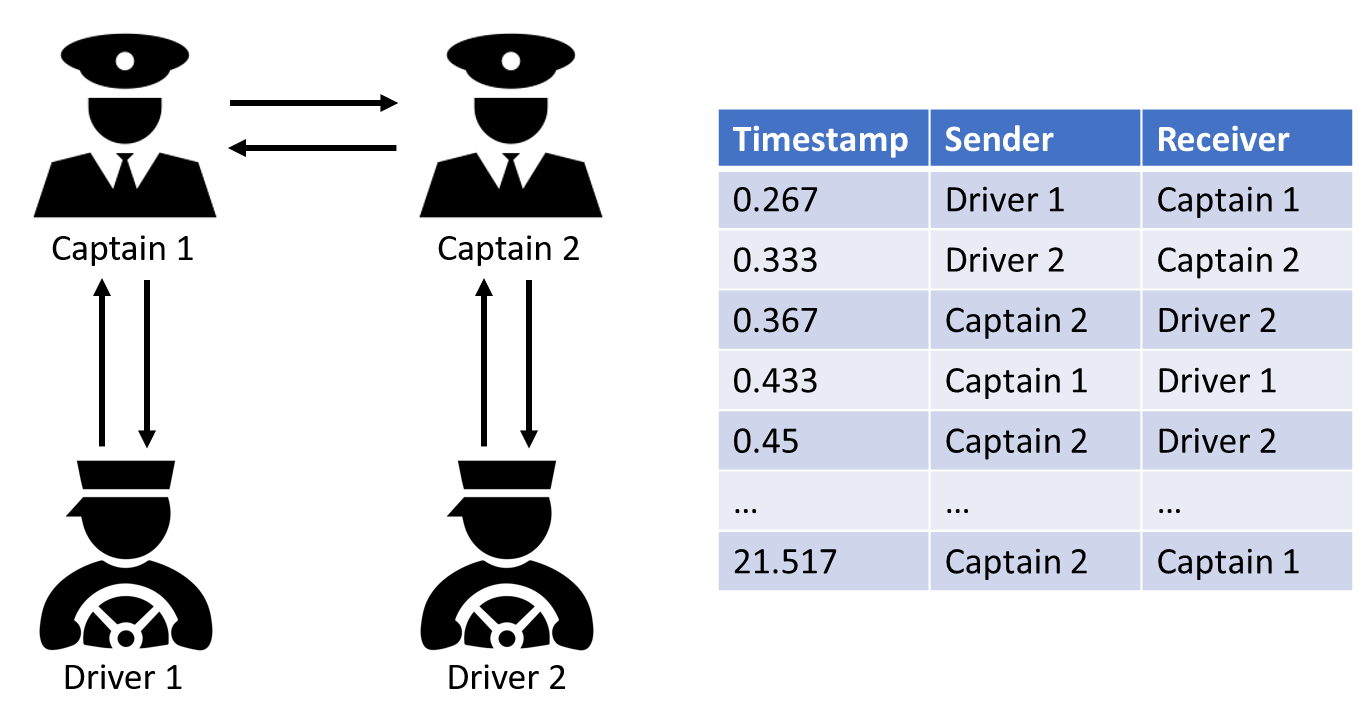}
    \caption{An illustration of the Multiple Team System data. The team structure and communication rules are shown on the left; communications are only permissible in directions indicated by the pointed arrows. The first and last rows of the dataset are shown on the right.}   
    \label{fig:mts}
\end{figure}

In this section we further elaborate on the distinction on agent identity discussed in Section \ref{subsec:unique_agent} through an empirical experiment demonstrating IRL's ability to characterize \textit{individual} behavioral preferences within a group interaction scenario, as opposed to the whole group perspective by the current REM. We show the kind of individual oriented insight IRL can learn in a group interaction event sequence modeling task.

\subsection{Data}
The data we use comes from a Multiple Team System (MTS) dataset collected and published by Pilny et al. \cite{pilny2014dynamic}. The dataset contains the sender, receiver, and timestamps of directed verbal messages (298 in total over a duration of 22 minutes) sent with Comm Net Radio (CNR) among 2 teams of 2 players each (a captain and a driver) playing a computer game where they carry out various virtual tasks to ensure the safety of a path for an emergency delivery, which requires substantial communication among the players. Within the game, a driver is only allowed to speak to its own captain whereas a captain can not only speak to its driver but also the other team's captain; however a captain cannot speak to the other team's driver. The communication scenario is illustrated in Figure \ref{fig:mts}. We choose the dataset for the following reasons: (1) it has been used to demonstrate REM modeling \cite{pilny2016illustration}, indicating that it is the typical type of data that's suitable for REM, and only if IRL is applied on data also suitable for REM can we highlight the additional insight IRL can provide; (2) with 4 actors and a singular type of directed social action (i.e. sending a message), the size of the dataset is relatively small, which is appropriate for our demonstrative purpose; (3) radio communication is a typical mode of technology-assisted social interaction especially used within team members that work together to achieve a goal such as emergency response and military missions, therefore the context of our demonstration has wider real-world applicability. 

\begin{figure}
    \centering
    \includegraphics[width = 0.8\columnwidth]{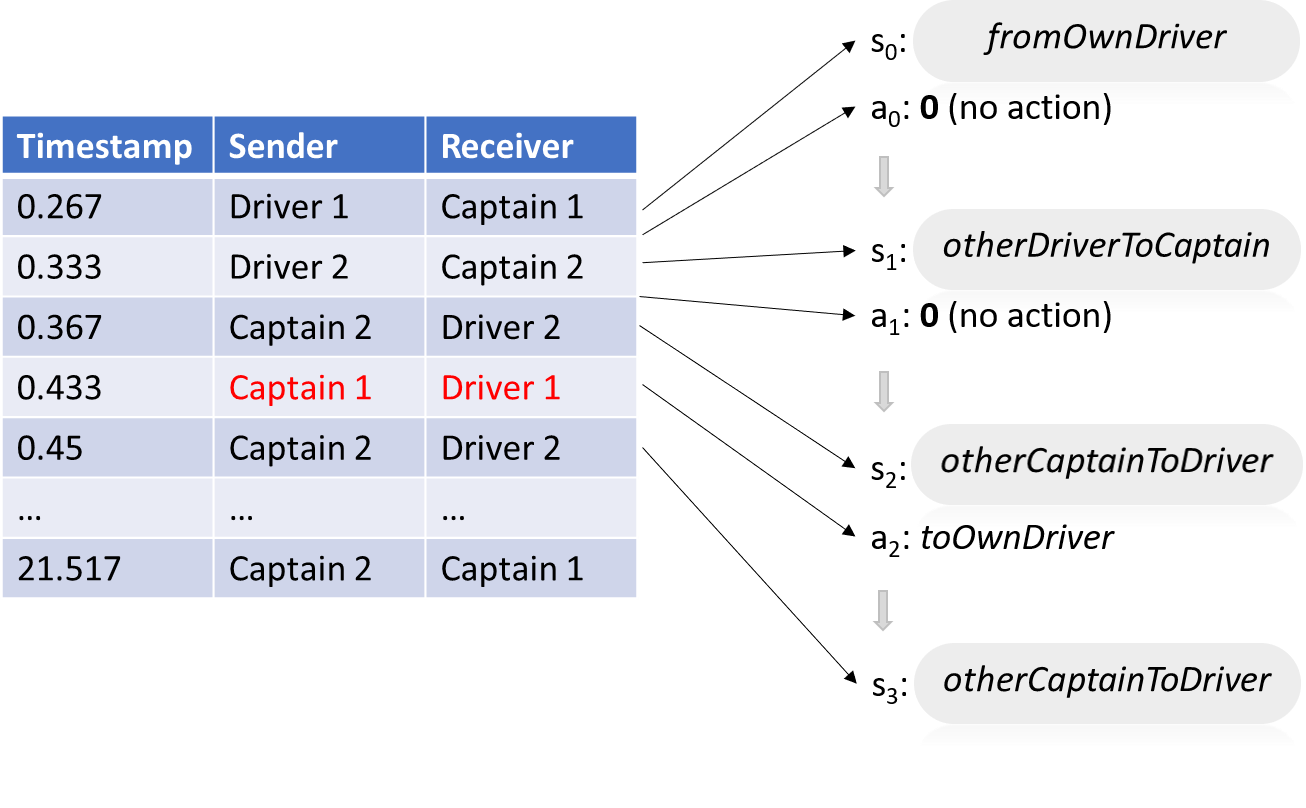}
    \caption{Converting the original timestamped event-list data (typical input data for REM, left) to a state-action trajectory from the perspective of Captain 1 for IRL use (right). Highlighted in red in the original data indicates an active action taken by the agent in question (Captain 1).}   
    \label{fig:mts_traj}
\end{figure}

\subsection{Experiment}
\begin{figure}[h]
    \centering
    \includegraphics[width = 0.8\columnwidth]{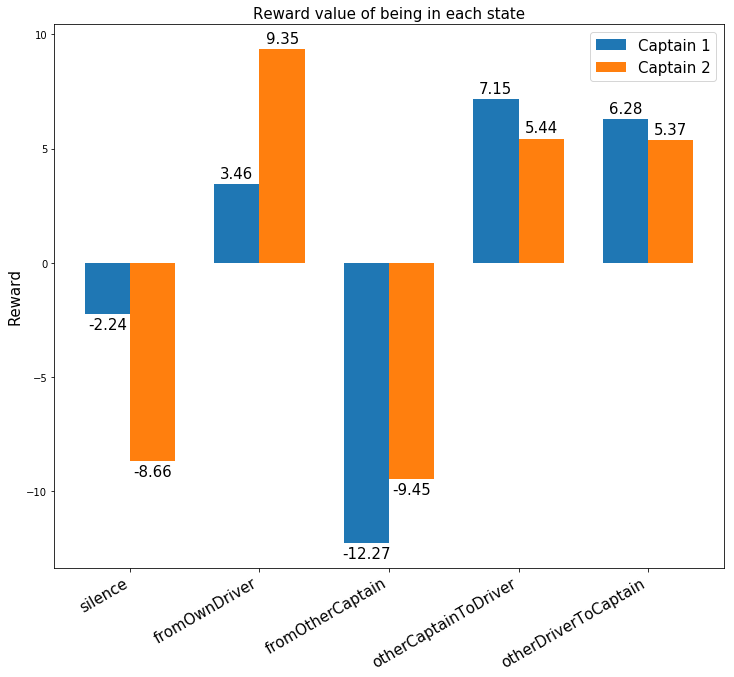}
    \caption{Reward values associated with each state for Captain 1 and Captain 2}   
    \label{fig:mts_rewards}
\end{figure}

To model individual behavioral preferences using IRL, we construct a simple Markov decision process as follows (among many possible, more complex ways). First we treat Captain 1 as the agent in question and consider all messages sent by Captain 1 as its actions and the rest as forming an environment upon which Captain 1 adjusts its behavior. Suppose that Captain 1 observes and re-evaluates the latest communication within the social system every time a new action is taken (by itself or other actors) and makes a decision as to what communication action to take next. From Captain 1's perspective there are 3 permissible actions: to do nothing ($\mathbf{0}$), to speak to its own driver ($toOwnDriver$), or to speak to the other team's captain ($toOtherCaptain$). Also from Captain 1's perspective, there are 5 possible states: \textit{silence} (manifested only when Captain 1 takes two actions back-to-back thus the state after the first action is that no other actors have ``chimed in" in between), getting spoken to by its own driver ($fromOwnDriver$), getting spoken to by the other team's captain ($fromOtherCaptain$), observing the other team's captain speaking to its own driver ($otherCaptainToDriver$), and observing the other team's driver speaking to its own captain ($otherDriverToCaptain$). Following these definitions, we re-adapt the original data into a state-action trajectory (illustrated in Figure \ref{fig:mts_traj}), compute state transition probabilities $P_{a}(s,s')$ empirically with add-one smoothing, and implement Maximum Entropy IRL algorithm \cite{ziebart2008maximum} \footnote{Adapted from the generic purpose python code available from Matthew Alger's Github repository \url{https://github.com/MatthewJA/Inverse-Reinforcement-Learning.git} \cite{alger16}} to learn a reward value associated with each of the 5 states. Then we switch perspective from Captain 1 to Captain 2, repeat the procedure described above, and learn an equivalent set of rewards for Captain 2. Due to symmetric, thus identical social structures from the respective perspective of Captains 1 and 2, we are able to compare their individual behavioral preferences based on the rewards values, which are shown in Figure \ref{fig:mts_rewards}. 

From the reward values we learned we observe that the two captains share similar preferences regarding communication. Both of them avoid the most being spoken to by the other team's captain (larger negative reward on $fromOtherCaptain$), moderately dislike back-to-back actions (smaller negative reward on $silence$), and enjoy being spoken to by its own driver and seeing chatters between members of the other team (positive reward). Differences exist as to how the desirability of different states compare for the two captains. Captain 1's most desirable scenario is to see conversations between members of the other team; however, Captain 2 places the highest reward on having received messages from its own driver. This distinction may reflect nuanced differences in the leadership style of the two captains. Such individual-oriented behavioral insights are the benefit of using IRL approach on group interaction data, whereas when using REM to model the same data, conclusions are drawn on group level tendencies such as ``group members interacted with members they trusted more" \cite{pilny2016illustration}. A limitation of our experiment is that we do not have ground truth data to validate the insights about individual behavioral preferences learned by inverse reinforcement learning, which we are motivated to address in future work. 

\section{Conclusion}

Identifying equivalences between different phenomena and mechanisms is a meaningful scientific endeavor that has greatly benefited the advancement of natural and applied sciences. James C. Maxwell introduced the mechanical-electrical analogies in the 19th century, revealing correspondences between velocity and current and between force and voltage. Such comparisons have later on facilitated deeper understanding and numerous novel developments in both domains of physics. We consider our effort presented in this paper of a similar nature and serving a similar purpose. We identified Inverse Reinforcement Learning as a novel approach to characterizing directed social interaction sequences in groups, which has typically been tackled by Relational Event Model previously. We examined the connections between REM and IRL, and found that both approaches posit a decision-making entity (group/agent) and a desirability quantity (rate/reward) for any particular state the entity is in, assume action choices based on the desirability of the potential state they lead the entity into, and optimize to solve for the desirability quantities. Some similarities are shared between REM and specific IRL algorithms such as the linear decomposition of the desirability quantity and the optimization procedure, while key distinctions exist in the model assumptions, state space specification, and model assumptions. Both approaches can be used to characterize sequences of directed social interaction events but produce different insights: using IRL for behavioral modeling in a group interaction is well-suited for learning \textit{individual} differences in behavioral preferences, which is difficult to accommodate with the REM approach.  

The applicability of REM and IRL extends beyond social interactions, as both essentially are theoretical frameworks of \textit{event sequence}. Social interaction dynamics can be considered as sequences of social action, and so can many tasks humans undertake. As a generalization of the REM framework, it has been used to extract sequential dependencies in participants' daily activities collected in the American Time Use Survey (ATUS) data \cite{marcum2015constructing}. On the IRL side, we found existing application in discovering underlying patterns in people's routine driving behavior \cite{banovic2016modeling}. With the growing amount of smart sensing data both in daily life and in workplaces such as manufacturing plant where task procedures are complex, we foresee great research opportunities in the application of REM and IRL for behavioral modeling. 

\bibliographystyle{plain}
\bibliography{main}

\end{document}